\pdfoutput=1
\documentclass[11pt]{article}

\usepackage[final]{acl}

\usepackage{times}
\usepackage{latexsym}

\usepackage[T1]{fontenc}
\usepackage[utf8]{inputenc}

\usepackage{microtype}

\usepackage{inconsolata}

\usepackage{graphicx}
\usepackage{soul}
\usepackage{amsmath}
\usepackage{amsfonts}
\usepackage{amsthm}
\usepackage{booktabs}
\usepackage{xspace}
\usepackage{subfigure}
\usepackage{pifont}
\usepackage{mathtools}
\usepackage{environ}
\usepackage{multirow}
\usepackage{pdfpages}
\usepackage{algorithm}
\usepackage{algorithmic}

\newcommand{\eg}{\emph{e.g.,}\xspace}

\newcommand{\ignore}[1]{}

\newcommand{\fullmodel}{\textbf{Med}ical dialogue system with knowledge \textbf{Ref}ining and dynamic prompt adjustment\xspace}
\newcommand{\model}{MedRef\xspace}

\title{Enhancing Medical Dialogue Generation through \\Knowledge Refinement and Dynamic Prompt Adjustment}

\author{
Hongda Sun$^{1*}$ \hspace{2mm} Jiaren Peng$^{2*}$ \hspace{2mm} Wenzhong Yang$^{3}$ \hspace{2mm} Liang He$^{4}$ \hspace{2mm} Bo Du$^{5}$ \hspace{2mm} Rui Yan$^{167\dagger}$ \\
{\normalsize
$^{1}$Gaoling School of Artificial Intelligence, Renmin University of China
} 
\hspace{2mm} 
{\normalsize
$^{2}$Sichuan University
}
\\
{\normalsize
$^{3}$School of Computer Science and Technology, Xinjiang University 
}
\hspace{2mm} 
{\normalsize
$^{4}$Tsinghua University
}
\\
{\normalsize
$^{5}$School of Computer Science, Wuhan University 
}
\hspace{2mm}
{\normalsize
$^{6}$School of Artificial Intelligence, Wuhan University 
}
\\
{\normalsize
$^{7}$Engineering Research Center of Next-Generation Intelligent Search and Recommendation, Ministry of Education 
}
\\
\texttt{\{sunhongda98, ruiyan\}@ruc.edu.cn, jiarenpeng666@gmail.com} \\
\texttt{yangwenzhong@xju.edu.cn, heliang@tsinghua.edu.cn, dubo@whu.edu.cn}
}

\begin{document}
\maketitle
\let\thefootnote\relax\footnotetext{$^\ast$Equal contribution.}
\let\thefootnote\relax\footnotetext{$^\dagger$Corresponding author: Rui Yan (\url{ruiyan@ruc.edu.cn}).}
\begin{abstract}
Medical dialogue systems (MDS) have emerged as crucial online platforms for enabling multi-turn, context-aware conversations with patients.
However, existing MDS often struggle to (1) identify relevant medical knowledge and (2) generate personalized, medically accurate responses. To address these challenges, we propose \model, a novel MDS that incorporates knowledge refining and dynamic prompt adjustment. First, we employ a knowledge refining mechanism to filter out irrelevant medical data, improving predictions of critical medical entities in responses. Additionally, we design a comprehensive prompt structure that incorporates historical details and evident details. To enable real-time adaptability to diverse patient conditions, we implement two key modules, Triplet Filter and Demo Selector, providing appropriate knowledge and demonstrations equipped in the system prompt.
Extensive experiments on MedDG and KaMed benchmarks show that MedRef outperforms state-of-the-art baselines in both generation quality and medical entity accuracy, underscoring its effectiveness and reliability for real-world healthcare applications.
\end{abstract}

\section{Introduction}
Medical dialogue systems (MDS) have emerged as a pivotal research spotlight, aiming to support healthcare professionals through multi-turn and context-aware conversations with patients~\cite{shi2024survey}.
Unlike general dialogue systems, MDS must understand and respond using medical domain knowledge~\cite{wei2018intro1,xu2019intro1,xia2020intro1}, offering valuable support for preliminary assessments and nursing care, particularly in resource-constrained environments~\cite{graham2014gap}.

\begin{figure}
    \centering
    \includegraphics[width=0.9\linewidth]{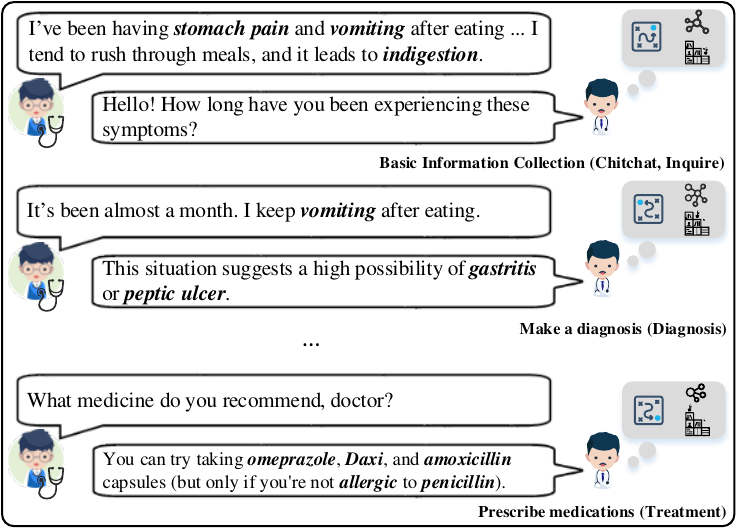}
    \caption{An example of medical dialogue generation.}
    \label{fig:intro}
\end{figure}

Despite the promise of MDS, several challenges remain in delivering accurate and contextually appropriate responses.
One key challenge is effectively tracking a patient's evolving health state throughout multi-turn interactions. 
As displayed in Figure~\ref{fig:intro}, doctors gradually refine their understanding of a patient’s condition over successive turns. Similarly, MDS must maintain coherence as the conversation progresses.
A common approach involves retrieving relevant medical entities (\eg symptoms, diagnoses, treatments) from a medical knowledge graph (MedKG)~\cite{vrbot,medpir}.
However, such retrieval-augmented generation (RAG) methods often introduce irrelevant knowledge, which degrades response quality.

Meanwhile, large language models (LLMs) have greatly improved MDS fluency, but remain sensitive to the prompt structure and content.
Effective prompts for MDS must (1) direct the model’s attention to critical medical entities and dialogue acts, and (2) include relevant conversation demonstrations for guidance.
Crucially, these prompts should dynamically adapt to reflect real-time patient information, which is underexplored by existing MDS.

To address these challenges, we aim to (1) refine the retrieved knowledge for more accurate response guidance and (2) dynamically adjust system prompts to align with specific patient conditions.
Therefore, we propose \model, a novel MDS with knowledge refining and dynamic prompt adjustment.
First, we explicitly represent the patient’s condition by incorporating contextual medical entities.
Inspired by~\cite{xu2023dfmed}, we adopt an entity-action joint prediction module to obtain the expected entities and acts.
To mitigate noise from retrieved entities, we introduce a knowledge refining mechanism to enable more accurate entity prediction and knowledge-driven response generation.
Building upon this, we construct a comprehensive prompt structure tailored to each dialogue turn. This system prompt mainly includes the following key components: \textbf{(1) Task instruction:} A high-level directive guiding the system's response generation process.
\textbf{(2) Historical details:} A summary of the dialogue context and identified medical entities.
\textbf{(3) Evident details:} Predicted entities and acts, and relevant knowledge triplets to provide medical evidence for response generation.
\textbf{(4) Relevant demonstration:} An example conversation for response formatting.
To enhance responsiveness, we integrate a dynamic prompt adjustment strategy that updates prompt contents in real time. 
Specifically, we leverage the Triplet Filter and Demo Selector to retain only the most relevant knowledge and demonstrations.
This enables our system to generate accurate, contextually grounded, and patient-specific responses throughout the dialogue.

We conduct extensive experiments on two widely used benchmarks: MedDG~\cite{liu2020meddg} and KaMed~\cite{vrbot}. Experimental results demonstrate the superiority of our \model compared with state-of-the-art baselines in both generation quality and medical entity accuracy. Ablation studies further validate the effectiveness of each module in our framework.

To sum up, our contributions can be summarized as follows:

$\bullet$ We propose \model, a novel medical dialogue system that jointly addresses knowledge redundancy and prompt adaptation for more accurate and context-aware response generation.

$\bullet$ We introduce a knowledge refining mechanism to filter out irrelevant information in retrieved knowledge, enhancing medical entity prediction and response grounding.

$\bullet$ We develop a dynamic prompt adjustment strategy that adapts prompt components in real time to the patient’s condition for improved personalization and coherence.

\section{Related Work}

\subsection{Medical Dialogue System}

Medical dialogue systems (MDS) are typically treated as a type of task-oriented dialogue system designed to assist in diagnosis and treatment~\cite{valizadeh2022task,varshney2022cdialog,sun2022debiased,sun2024collaborative}.
However, progress in this area is often limited in collecting large-scale medical datasets due to privacy and ethical concerns. 
To address this, \citet{zeng2020meddialog} released MedDialog, a large-scale Chinese-English medical dialogue dataset, which features a larger number of conversation sessions with relatively short turns.
\citet{liu2020meddg} introduced MedDG with medical entity annotations in each utterance, facilitating more fine-grained analysis.
Early studies on MDS rely on template-based methods for various tasks like information extraction~\cite{peng2024one,zhang2020extractor}, relation prediction~\cite{du2019relations,lin2019relation,xia2021relation}, and slot filling~\cite{shi2020slot}.
More recently, response generation has gained focus, leveraging sequence-to-sequence models~\cite{bahdanau2014seq2seq,vaswani2017attention,see2017pointer} and pre-trained models like BioBERT~\cite{lee2020biobert}, MedBERT~\cite{rasmy2021medbert}, GPT-2~\cite{radford2019gpt2}, and DialoGPT~\cite{zhang2019dialogpt}.
MDS require integration of medical knowledge for accurate responses.
Building upon this,
VRBot~\cite{vrbot} formulates patient states and physician actions for response generation.
MedPIR~\cite{medpir} recalls pivotal information as a prefix to generate responses.
DFMed~\cite{xu2023dfmed} uses a dual flow enhanced framework to sequentially model the medical entities and dialogue acts.

\subsection{\mbox{Knowledge-Grounded Dialogue Generation}}
Knowledge-grounded conversations (KGC) aim to generate responses based on background knowledge retrieved from knowledge graphs~\cite{speer2017kgc,ghazvininejad2018kgc,li2020kgc,chen2020kgc}.
The background knowledge is generally retrieved from structured and unstructured sources. 
The unstructured knowledge used in KGC is mainly documents or paragraphs~\cite{dinan2018unstructured,zhang2018unstructured,kim2020unstructured,zhao2020unstructured}. 
Structured KGC, on the other hand, relies on knowledge triplets or graphs to predict key entities~\cite{liu2018structured,tuan2019structureddykg,xu2020structured}.
Given the dependence of medical dialogues on domain-specific knowledge, KGC methods have been widely applied using medical knowledge graphs (MedKG) to support informed responses~\cite{vrbot,medpir}.

However, existing approaches often retrieve irrelevant information from MedKG, misaligning with a patient's specific condition. Therefore, we propose a knowledge refining mechanism for improved entity prediction and response generation.

\section{Method}

\begin{figure*}[ht]
    \centering
    \includegraphics[width=0.9\textwidth]{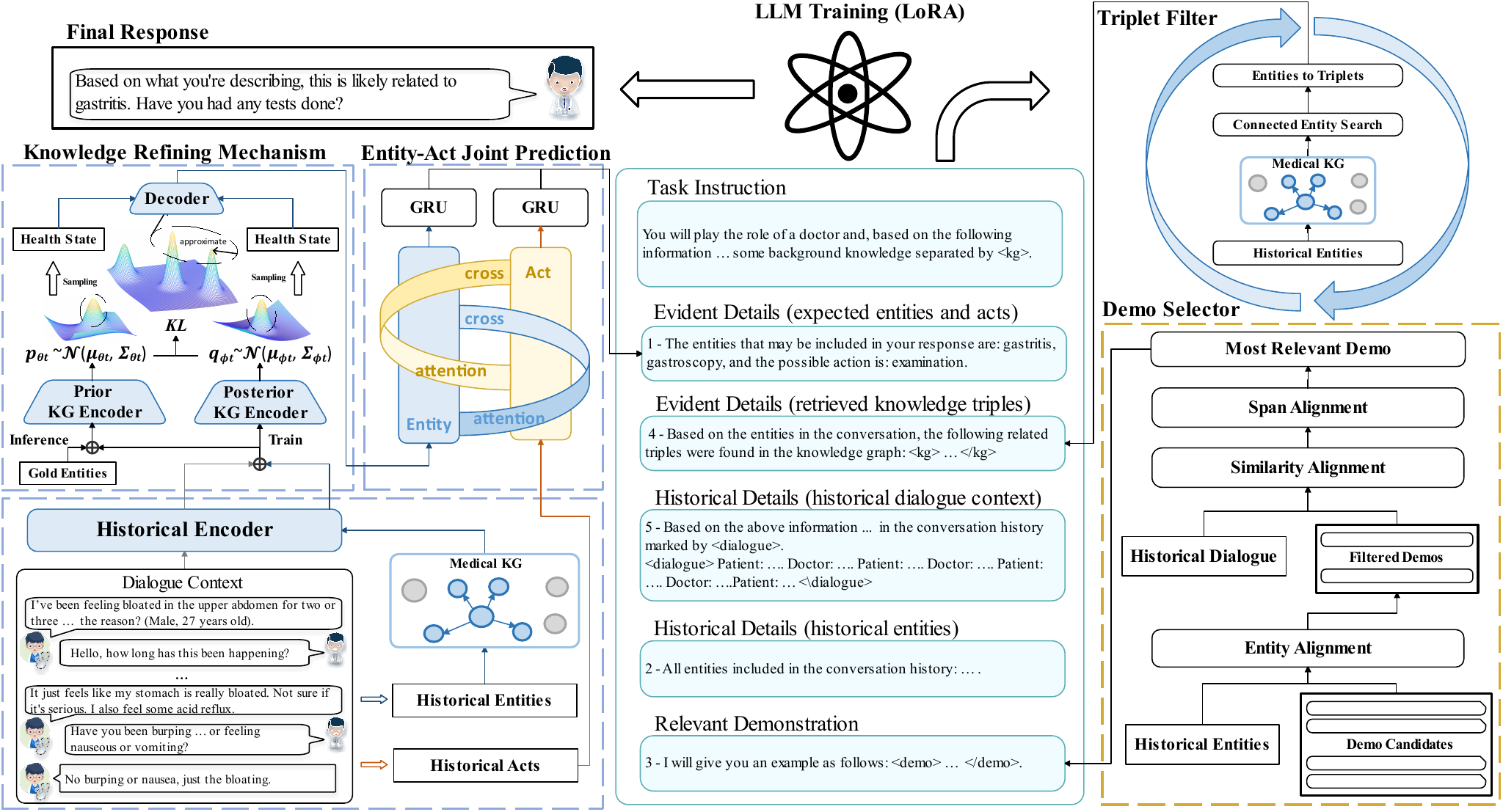}
    \caption{System overview of our \model, involving encoding dialogue history, refining retrieved knowledge, and jointly predicting entities and acts. The triplet filter and demo selector are used to enhance the prompt for final response generation.}
    \label{fig:main}
\end{figure*}

\subsection{Problem Formulation}
Suppose a medical conversation session $c = \{ u_1, r_1, u_2, r_2, \dots, u_T, r_T \}$ lasts for a total of $T$ turns of utterances, where $u_t$ and $r_t$ represent the patient’s utterance and the doctor’s response at the $t$-th turn. 
The dialogue context at each turn $t$ is denoted as $\overline{c}_t = \{ u_1, r_1, \dots, u_{t-1}, r_{t-1}, u_t \}$, which conditions the generation of the current doctor response $r_t$.
Each utterance introduces multiple medical entities, and each doctor's response is further annotated with dialogue acts.
The historical medical entities $\overline{x}_t$ and dialogue acts $\overline{a}_t$ within $\overline{c}_t$ guide the generation of the response $r_t$.
Moreover, a medical knowledge graph $G$ is commonly used to retrieve relevant knowledge to aid in response generation. Therefore, the objective of MDS is to generate the doctor response $r_t$ at each turn $t$, conditioned on the dialogue context $\overline{c}_t$, historical entities $\overline{x}_t$, historical acts $\overline{a}_t$ and relevant knowledge from $G$.

\subsection{Input Representation}
To effectively track the patient’s health condition and generate appropriate responses, it is essential to encode the key components of the dialogue history in the MDS.
In the context $\overline{c}_t$, each patient utterance is denoted as $u_i = (u_{i,1}, u_{i,2}, \cdots, u_{i,|U_i|})$ with $|U_i|$ tokens, and each doctor utterance as $(r_{j,1}, r_{j,2}, \cdots, r_{j,|R_j|})$ with $|R_j|$ tokens. To capture their semantic content, we first apply an embedding layer $f_{emb}$, yielding token-level embeddings $e_{u_i}$ and $e_{r_j}$ for patient and doctor utterances, respectively.
Given the medical nature of the task, we adopt MedBERT~\footnote{https://github.com/trueto/medbert}, a pre-trained model specialized in medical domains, as our encoder backbone. 
The embedded utterances are processed by this encoder $f_{enc}$ to incorporate sequential dialogue information, and the final output $e_{\overline{c}_t}$ serves as the contextual representation for subsequent modules. 
The encoding process can be formalized:
\begin{equation}
\begin{gathered}
    e_{u_i} = f_{emb}(u_{i,1}, u_{i,2}, \cdots, u_{i,|U_i|}), \\
    e_{r_j} = f_{emb}(r_{j,1}, r_{j,2}, \cdots, r_{j,|R_j|}), \\
    e_{\overline{c}_t} = f_{enc}(e_{u_1}, e_{r_1},\cdots,e_{u_t}).
\end{gathered}
\end{equation}

We then retrieve related entities from a medical knowledge graph $G$ to guide accurate response generation.
Specifically, we construct a subgraph $G_{\overline{x}_t}^0 = \{G_{\overline{x}_{t_1}}^0, \ldots, G_{\overline{x}_{t_m}}^0\}$ that totally contains $m$ historical entities $\overline{x}_t$ and their one-hop neighbors. Then we encode these entities using $f_{enc}$ and structural information via a graph attention network (GAT)~\cite{gat}, $f_{gat}$. This yields the subgraph representation:
\begin{equation}
    e_{\overline{x}_t}^{G_0} = f_{gat} ( f_{enc}(G_{\overline{x}_{t_1}}^0, \ldots, G_{\overline{x}_{t_m}}^0)).
    \label{eq:G1}
\end{equation}

Moreover, the dialogue acts capture the communicative intents of each response (\eg symptom inquiry, disease diagnosis, and treatment suggestion). The historical dialogue acts are encoded as act-level representations $e_{\overline{a}_t}$.
These enriched representations collectively provide the contextual information for accurate response generation.

\subsection{Knowledge Refining Mechanism}

Retrieved entities can be noisy or overly broad due to deterministic retrieval.
To address this, we use a knowledge refinement mechanism that models a latent variable $z_t$ to filter irrelevant knowledge.
We first estimate the prior distribution $p_\theta(z_t|\overline{c}_t, G_{\overline{x}_t}^0)$ based on the dialogue context $\overline{c}_t$ and retrieved entities $G_{\overline{x}_t}^0$. To guide the prior toward retaining useful knowledge, we define a posterior distribution $q_\phi(z_t|\overline{c}_t, G_{\overline{x}_t}^0, x_t)$ by incorporating the ground-truth entities $x_t$ from the target response $r_t$.
Both prior and posterior are modeled as Gaussian distributions and parameterized via separate encoders:

{\fontsize{7.8}{0}\selectfont
\begin{equation}
\begin{gathered}
    p_\theta(z_t|\overline{c}_t, G_{\overline{x}_t}^0) = \mathcal{N}(\mu_\theta(e_{\overline{c}_t}, e^{G_0}_{\overline{x}_t}), \Sigma_\theta(\overline{c}_t, e^{G_0}_{\overline{x}_t})), \\
    q_\phi(z_t|\overline{c}_t, G_{\overline{x}_t}^0, x_t) = \mathcal{N}(\mu_\phi(\overline{c}_t, e^{G_0}_{\overline{x}_t}, x_t), \Sigma_\phi(\overline{c}_t, e^{G_0}_{\overline{x}_t}, x_t)),
\end{gathered}
\end{equation}
}
where $\mu_\theta$, $\Sigma_\theta$, $\mu_\phi$, and $\Sigma_\phi$ are computed from separate knowledge encoder networks.
Once the latent factor $z_t$ is sampled, it is passed through the knowledge decoder $f_{dec}$, and its output is combined with the original entity embedding $e_{\overline{x}_t}^{G_0}$ to produce the refined representation:
\begin{equation}
e_{\overline{x}_t}^{G} = f_{dec}(z_t) + e_{\overline{x}_t}^{G_0}
\end{equation}
This refined embedding $e_{\overline{x}_t}^{G}$, with reduced noise and improved relevance, is used to better predict the expected entities in the response.

\subsection{Entity-Act Joint Prediction}
Based on the refined knowledge, we can reconstruct the entities in the response.
To capture the high correspondence between medical entities (symptoms, diseases, and treatments) and dialogue acts (symptom inquiry, disease diagnosis, and treatment suggestions), we leverage a joint prediction module to obtain the expected entities and acts in the target response.
We first model interactions between context, refined entities, and historical acts using a cross-attention module $f_{ca}$, followed by a GRU $f_{gru}$ to obtain new representations:
\begin{equation}
\begin{gathered}
        e_{t}^{CG} = f_{ca}(e_{\overline{c}_t}, e_{\overline{x}_t}^{G}), \\
        e_{t}^{CGA} = f_{ca}(e_{t}^{CG}, e_{\overline{a}_t}), \\
        \widetilde{e}_{\overline{x}_t}^G = f_{gru}(e_{\overline{x}_t}^{G} \oplus e_t^{CG} \oplus e_t^{CGA}). 
\end{gathered}
\label{eq:equation1}
\end{equation}
\begin{equation}
\begin{gathered}
        e_t^{CA} = f_{ca}(e_{\overline{c}_t}, e_{\overline{a}_t}), \\
        e_t^{CAG} = f_{ca}(e_t^{CA}, e_{\overline{x}_t}^G), \\
        \widetilde{e}_{\overline{a}_t} = f_{gru}(e_{\overline{a}_t} \oplus e_t^{CA} \oplus e_t^{CAG}). 
\end{gathered}
\label{eq:equation2}
\end{equation}
We then compute prediction probabilities for entities and acts in the $t$-th turn via linear transformation layers along with sigmoid $\sigma(\cdot)$ as activation functions.
\begin{equation}
\begin{gathered}
        \widehat{x}_t = \sigma(W_x \widetilde{e}_{\overline{x}_t}^G + b_x), \\
        \widehat{a}_t = \sigma(W_a \widetilde{e}_{\overline{a}_t} + b_a),
\end{gathered}
\label{eq:equation3}
\end{equation}
where $W_x \in \mathbb{R}^{|X| \times d}$ and $b_x \in \mathbb{R}^{|X|}$; $W_a \in \mathbb{R}^{|A| \times d}$ and $b_a \in \mathbb{R}^{|A|}$. $|X|$ and $|A|$ are the numbers of candidate entities and acts, and $d$ is the hidden size.

\subsection{Dynamic Prompt Adjustment}

\subsubsection{Prompt Design}

To better motivate LLMs to generate accurate and patient-specific responses, we design a comprehensive prompt structure. 
As shown in Figure~\ref{fig:main}, the system prompt $\mathcal P = [\mathcal{I}; \mathcal{H}; \mathcal{K}; \mathcal{E}]$ contains the following key components:

\textbf{Task instruction} $\mathcal I$ outlines the task that responds to the patient and explains the structure of the remaining prompts.
\textbf{Historical details} $\mathcal H$ summarize key elements in the dialogue history, including the dialogue context $\overline{c}_t$, and sequentially listed historical entities $\overline{x}_t$ and acts $\overline{a}_t$.
\textbf{Evident details} $\mathcal{K}$ provide medical knowledge for generating responses, containing predicted entities and acts, and relevant knowledge triplets from MedKG.
\textbf{Relevant demonstration} $\mathcal E$ provides an in-context example to guide response formatting.

To enable real-time adaptation for varying patient conditions, we integrate a dynamic prompt adjustment strategy by introducing the Triplet Filter and Demo Selector modules to refine the equipped knowledge and demonstrations in the prompt.

\subsubsection{Triplet Filter}
To obtain reliable knowledge triplets from retrieved entities, we design an iterative filtering process.

First, the retrieved one-hop subgraph $G_{\overline{x}_t}^0$  is transformed into a set of triplets $Tri_{\overline{x}_t}^0$. Next, we compute the frequency of each entity in these triplets and sort them in descending order. Based on these frequencies, we dynamically adjust the triplets retained by setting a threshold $\tau$. Those triplets can be kept if and only if their head and tail entities both have frequencies not less than $\tau$.

{\fontsize{8.5}{0}\selectfont
\begin{equation}
    Tri_{\overline{x}_t}^\tau = \{(e_{head}, r, e_{tail})| \min(\# e_{head}, \# e_{tail}) \ge \tau \}
\end{equation}
}

Initially, $\tau$ is set to 1 and is incremented in each iteration, gradually reducing the number of retained triplets. The process terminates once the number of triplets in $Tri_{\overline{x}_t}^\tau$ does not exceed a predefined maximum $M$. The current $Tri_{\overline{x}_t}^\tau$ is then used as part of the final evident details in the prompt.

\subsubsection{Demo Selector}

To select the most relevant demonstration for the system prompt, we introduce a multi-step alignment process.

\paragraph{Entity alignment.}
We begin by organizing all training conversations into subsets based on entity annotations in the first patient utterance. Specifically, we construct multi-entity subsets $S_E = \{S_{E_1}, \ldots, S_{E_K}\}$, where each subset $S_{E_k}$ contains conversation cases whose first utterance includes the same $n$ entities $E_k = \{x_1, \ldots, x_n\}$. In parallel, we create single-entity subsets $S_e$, where each subset $S_{e'}$ contains cases with first utterances that mention the shared entity $e'$.

Given a current dialogue context $\overline{c}_t$, we need to check whether its first utterance $u_1$ exactly matches any entity set in $S_E$. If so, we retrieve the corresponding subset as the candidate demo set $S_{demo}$. Otherwise, we fall back to the single-entity subsets and select all sessions from $S_e$ that share at least one entity with 
$u_1$.

\paragraph{Similarity alignment.}
To refine the demo selection, we compute the semantic similarity between the current first utterance $u_1$ and those in $S_{demo}$. We encode each candidate separately and then apply cosine similarity to identify the closest conversation $c_{full}$ as the demonstration reference.

\paragraph{Span alignment.}
To improve contextual relevance and reduce prompt length, we extract a focused span from $c_{full}$ using a sliding window of size $\xi$. Let the total utterance sequence of $c_{full}$ be $\{u_1, r_1, u_2, r_2, \dots, u_T, r_T\}$, and denote the start index as $i_s = 2t-1$, corresponding to the current dialogue turn $t$. The final demonstration $\mathcal{E}$ is intercepted in three cases: (1) If $i_s\leq \xi$, we select the first $ 2\xi $ utterances from $c_{full}$; (2) If $\xi<i_s<T-\xi$, we select the utterances from index $ i_s - \xi $ to $ i_s + \xi $;
(3) If $T-\xi\leq i_s$, we select the last $ 2\xi $ utterances.

\subsection{Model Optimization}
To optimize different modules of \model, we design a two-stage training objective.
We first pre-train the entity-act joint prediction module in preparation for subsequent response generation. 
For predicting medical entities, we compute the binary cross-entropy (BCE) loss $\mathcal{L}_{x}$ between predictions $\widehat{x}_t$ and ground-truth entity labels $x_t$. Similarly, dialogue act prediction is trained based on the cross-entropy loss $\mathcal{L}_{a}$. These loss functions can be formulated as:

{\fontsize{8.5}{0}\selectfont
\begin{equation}
\begin{gathered}
        \mathcal{L}_{x}  = - \sum_{t=1}^{T} \sum_{i=1}^{|X|}[ x_{t_i}\log(\widehat{x}_{t_i}) + (1-x_{t_i}) \log(1-\widehat{x}_{t_i})], \\
        \mathcal{L}_{a}  = - \sum_{t=1}^{T} \sum_{j=1}^{|A|}[ a_{t_j}\log(\widehat{a}_{t_j}) + (1-a_{t_j}) \log(1-\widehat{a}_{t_j})],
\end{gathered}
\label{eq:equation4}
\end{equation}
}

To ensure consistency in knowledge refining, we minimize the Kullback-Leibler (KL) divergence between the prior $p_\theta$ and posterior $q_\phi$:

{\fontsize{9.5}{0}\selectfont
\begin{equation}
    \mathcal{L}_{kl} = \sum_{t=1}^{T} D_{KL}(q_\phi(z_t|\mu_{\phi}, \Sigma_{\phi_t}) || p_\theta(z_t|\mu_{\theta}, \Sigma_{\theta_t})).
\end{equation}
}

We assign the weights $\lambda_x$, $\lambda_a$, and $\lambda_{kl}$ to each loss, and the overall loss function for this stage is a weighted combination:
\begin{equation}
        \mathcal{L}=\lambda_x\mathcal{L}_x+\lambda_a\mathcal{L}_a+\lambda_{kl}\mathcal{L}_{kl}.
    \label{eq:G5}
\end{equation}

Next, with the prediction module fixed, we fine-tune the medical LLM responsible for response generation.
By maximizing the log-likelihood of the system responses, the language model based loss is given by:
\begin{equation}
    \mathcal{L}_{gen} = -\sum_{t=1}^T \log \sum_{k} p_{gen}(r_{t_k}|r_{t_{<k}}, \mathcal{P}).
\end{equation}

\begin{table*}[t]\small
\centering
\caption{Comparison results on MedDG and KaMed datasets. ``B''=BLEU, ``R''=ROUGE, ``E-F1''=entity-F1. Bold/underline numbers denote significant improvements ($p$-value<0.01) over the second-best.}
\label{CombinedResults}
\setlength{\tabcolsep}{4pt}
\setlength{\abovecaptionskip}{-5pt}
\renewcommand{\arraystretch}{1.2}
\begin{tabular}{@{}llcccccc|cccccc@{}}
\toprule
\multirow{2}{*}{\textbf{Category}} & \multirow{2}{*}{\textbf{Method}} & \multicolumn{6}{c}{\textit{MedDG}} & \multicolumn{6}{c}{\textit{KaMed}} \\
\cmidrule(lr){3-8} \cmidrule(lr){9-14}
& & \textbf{B-1} & \textbf{B-2} & \textbf{B-4} & \textbf{E-F1} & \textbf{R-1}& \textbf{R-2} & \textbf{B-1} & \textbf{B-2}& \textbf{B-4} &\textbf{E-F1} & \textbf{R-1} &\textbf{R-2} \\
\midrule
\multirow{2}{*}{DL-based} 
& Seq2Seq & 28.55 & 22.85 & 15.45 & 12.88 & 25.61 & 11.24 & 23.52 & 18.56 & 12.13 & - & 23.56 & 8.67 \\
& VRBot & 29.69 & 23.90 & 16.34 & 12.78 & 24.69 & 11.23 & 30.04 & 23.76 & 16.36 & 12.08 & 18.71 & 7.28 \\ \hline
\multirow{3}{*}{PLM-based}
& GPT-2 & 35.27 & 28.19 & 19.16 & 16.14 & 28.74 & 13.61 & 33.76 & 26.58 & 17.82 & 17.26 & 26.80 & 10.56 \\
& BART & 34.94 & 27.99 & 19.06 & 16.66 & 29.03 & \underline{14.40} & 33.62 & 26.43 & 17.64 & 19.20 & 27.91 & 11.43 \\
& DFMed & 41.74 & \underline{32.93} & 22.48 & \underline{21.54} & 28.90 & 13.71 & 39.59 & 30.53 & 20.30 & \underline{21.33} & 27.67 & 11.21 \\ \hline
\multirow{5}{*}{LLM-based}
& DISC-MedLLM & 40.72 & - & 22.60 & 10.15 & 20.13 & 6.6 & 38.05 & - & 20.26 & 13.54 & 20.48 & 5.93 \\
& GPT-4o & \underline{42.19} & - & \textbf{23.32} & 13.15 & 13.99 & 3.47 & \textbf{41.88} & - & \underline{23.34} & 13.86 & 13.94 & 3.1 \\
& HuatuoGPT-II & 39.03 & 32.56 & 23.02 & 8.67 & 10.94 & 1.76 & 40.35 & \textbf{32.93} & \textbf{23.92} & 12.00 & 13.84 & 2.74 \\
& Zhongjing & 26.65 & 21.75 & 15.02 & 6.43 & 13.14 & 2.82 & 27.48 & 22.35 & 15.52 & 6.44 & 13.70 & 3.05 \\
& Chatglm3-6B & 33.16 & 26.51 & 17.97 & 17.43 & \underline{29.27} & 13.69 & 32.03 & 25.20 & 16.68 & 20.56 & \underline{28.02} & \underline{12.12} \\ \hline
& \textbf{\model} & \textbf{43.51} & \textbf{33.82} & \underline{23.04} & \textbf{22.70} & \textbf{30.07} & \textbf{14.52} & \underline{40.47} & \underline{31.62} & 21.28 & \textbf{21.96} & \textbf{28.14} & \textbf{12.42} \\ 
\bottomrule
\end{tabular}
\end{table*}

\section{Experimental Setup}

\subsection{Datasets}

We conduct experimental evaluations on two widely used benchmarks, MedDG and Kamed. MedDG contains over 17,000 medical dialogues annotated with 160 medical entities across 5 categories: diseases, symptoms, medications, examinations, and attributes. It is officially split into 14,862 (train), 1,999 (validation), and 999 (test) sessions. Kamed includes over 63,000 dialogues spanning 100+ departments. Following DFMed~\cite{xu2023dfmed}, we remove privacy-sensitive data, resulting in 29,159 (train), 1,532 (validation), and 1,539 (test) sessions.
Dialogue acts are labeled into 7 types: \textit{Chitchat, Inform, Inquire, Provide Daily Precaution, State a Required Medical Test, Make a Diagnosis, and Prescribe Medications.}

\subsection{Baselines}
We compare \model against the following three types of baselines: (1)
\textbf{DL-based methods}:
Seq2Seq~\cite{sutskever2014sequence}, RNN with attention;
VRBOT~\cite{vrbot}, patient state and physician action tracking model. (2) \textbf{PLM-based methods}:
GPT-2~\cite{radford2019gpt2},
BART~\cite{lewis2019bart}, general-purpose generative models;
DFMed~\cite{xu2023dfmed}, dual flow model leveraging interwoven entities and acts.
(3) \textbf{LLM-based methods}:
Chatglm3-6B~\cite{du2022glm}, general LLM fine-tuned on medical dialogues;
Zhongjing~\cite{yang2024zhongjing}, traditional Chinese medicine dialogue model;
HuatuoGPT-II~\cite{chen2023huatuogpt} (Baichuan-7B), DISC-MedLLM~\cite{bao2023disc} (Baichuan-13B), specialized medical LLM;
GPT-4o~\cite{gpt4o}, advanced closed-source LLM.

\subsection{Evaluation Metrics}

\paragraph{Automatic evaluation.}
To evaluate the quality of the model's generated responses, we utilize \textbf{BLEU}~\cite{papineni2002bleu} and \textbf{ROUGE}~\cite{lin2004rouge} for assessing lexical similarity, and \textbf{entity-F1} score to measure entity-level accuracy.

\paragraph{Human evaluation.}
We focus on three key human evaluation metrics: \textbf{fluency (FLU)} measures how naturally and smoothly the conversation flows; \textbf{knowledge accuracy (KC)} focuses on the correctness of the medical terms; and \textbf{overall quality (OQ)} considers the holistic response effectiveness.

\subsection{Implementation Details}
\label{sec:app-implement-details}
We use ChatGLM3-6B as the backbone of our response generator, which is fine-tuned 
with LoRA~\cite{hu2021lora} (rank=8, $\alpha$=32, dropout=0.1) using AdamW (lr=5e-5).
MedBERT~\cite{rasmy2021medbert} is used for entity and act prediction (lr=3e-5, batch size=8).
We retrieve up to $M$=25 triplets from CMeKG~\cite{cmekg}. The sliding window size is $\xi$=2. The loss weights are set to $\lambda_x=1$, $\lambda_a=0.05$, $\lambda_{kl}=0.05$.$^1$\footnote{$^1$Our code is available at \url{https://github.com/simon-p-j-r/MedReF}.}

\begin{table*}[t]
\setlength{\tabcolsep}{4pt}
\renewcommand{\arraystretch}{1.1}
\centering
\caption{Ablation results of \model on MedDG and KaMed datasets.}
\label{CombinedAblation}
\begin{tabular}{@{}lcccccc|cccccc@{}}
\toprule
\multirow{2}{*}{\textbf{Method}} & \multicolumn{6}{c}{\textit{MedDG}} & \multicolumn{6}{c}{\textit{KaMed}} \\
\cmidrule(lr){2-7} \cmidrule(lr){8-13}
& \textbf{B-1} & \textbf{B-2} & \textbf{B-4} & \textbf{E-F1} & \textbf{R-1}& \textbf{R-2} & \textbf{B-1} & \textbf{B-2}& \textbf{B-4} &\textbf{E-F1} & \textbf{R-1} &\textbf{R-2} \\
\midrule
\textbf{\model} & \textbf{43.51} & \textbf{33.82} & \textbf{23.04} & \textbf{22.70} & \textbf{30.07} & \textbf{14.52} & \textbf{40.47} & \textbf{31.62} & \textbf{21.28} & \textbf{21.96} & \textbf{28.14} & \textbf{12.42} \\ 
\midrule
w/o KRM & \underline{42.58} & \underline{33.45} & \underline{22.70} & \underline{21.94} & \underline{29.88} & \underline{14.23} & \underline{40.29} & \underline{31.10} & \underline{20.88} & \underline{21.51} & 27.95 & 11.92 \\
w/o Demo & 41.80 & 32.87 & 22.31 & 21.84 & 29.69 & 13.93 & 39.07 & 30.34 & 20.46 & 20.09 & 27.35 & 11.90 \\
w/o Kg & 41.76 & 32.83 & 22.24 & 21.58 & 29.86 & 13.93 & 39.82 & 30.96 & 20.81 & 20.55 & \underline{28.09} & 11.87 \\
E-A\&Cxt only & 41.63 & 32.75 & 22.30 & 21.30 & 28.68 & 13.27 & 39.30 & 30.38 & 20.42 & 20.81 & 26.72 & 11.22 \\
Cxt only & 33.16 & 26.51 & 17.97 & 17.43 & 29.27 & 13.69 & 32.03 & 25.20 & 16.68 & 20.56 & 28.02 & \underline{12.12} \\
\bottomrule
\end{tabular}
\end{table*}

\section{Experimental Results}
\subsection{Overall Performance}

As shown in Table \ref{CombinedResults}, \model consistently outperforms all baselines across multiple metrics, demonstrating its effectiveness in generating high-quality, medically grounded responses.
Compared to GPT-4o, \model achieves +1.32\% BLEU-1, +16.08\% ROUGE-1, and +11.05\% Entity-F1, demonstrating superior lexical alignment, fluency, and medical accuracy. This advantage stems from task-specific fine-tuning, whereas GPT-4o’s closed-source nature limits its adaptability to medical dialogue nuances.
\model tends to generate fluent utterances that align well with human-authored responses, contributing to its superior ROUGE and entity-F1 scores, reflecting content richness and relevance.

However, \model slightly underperforms HuatuoGPT-II and GPT-4o on BLEU scores on KaMed. This discrepancy may arise from the dataset complexity and the response style bias of these models. First, KaMed spans a broader range of clinical scenarios, encompassing over 100 departments, which increases the complexity of the required medical knowledge and makes learning high-coverage representations more challenging. Besides, HuatuoGPT-II and GPT-4o often generate verbose, QA-style replies. While this verbosity can increase token-level overlap with references (thereby inflating BLEU scores), it tends to introduce irrelevant or redundant content, leading to much lower entity-F1 scores.
Second, HuatuoGPT-II and GPT-4o tend to adopt a QA-style approach to addressing patient inquiries, often generating very long text responses with redundancy and nonsense. This response trend is not enough to the point that it helps to slightly improve the BLEU indicator, but significantly reduces the entity F1 score. 

\subsection{Ablation Study}
To investigate the contribution of each module in the proposed system, we conduct a comprehensive ablation study that includes the following variants for comparison:
(1) \textbf{w/o KRM} removes knowledge refinement mechanism.
(2) \textbf{w/o Demo} removes the demonstration $\mathcal{E}$ matched by the demo selector.
(3) \textbf{w/o Kg} removes the knowledge triplets retrieved from MedKG.
(4) \textbf{E-A\&Cxt only} retains only the predicted entities and actions along with the dialogue context; no demonstrations or external knowledge are provided, and the KRM is not used.
(5) \textbf{Cxt only} uses only the dialogue context, without any additional guidance or knowledge.

The ablation results in Table \ref{CombinedAblation} show that all variant models exhibit noticeable performance declines, underscoring the importance of each component.
In particular, w/o KRM suffers the most significant drop across all evaluation metrics, highlighting its dual role in filtering out redundant knowledge and improving entity prediction accuracy.
Moreover, the performance degradation of other model variants relative to the full model illustrates the importance of prompt integrity and also shows that the retrieval knowledge and demonstrations selected into our prompt are more relevant than before.

\begin{figure*}[t]
    \centering
    \includegraphics[width=0.8\textwidth]{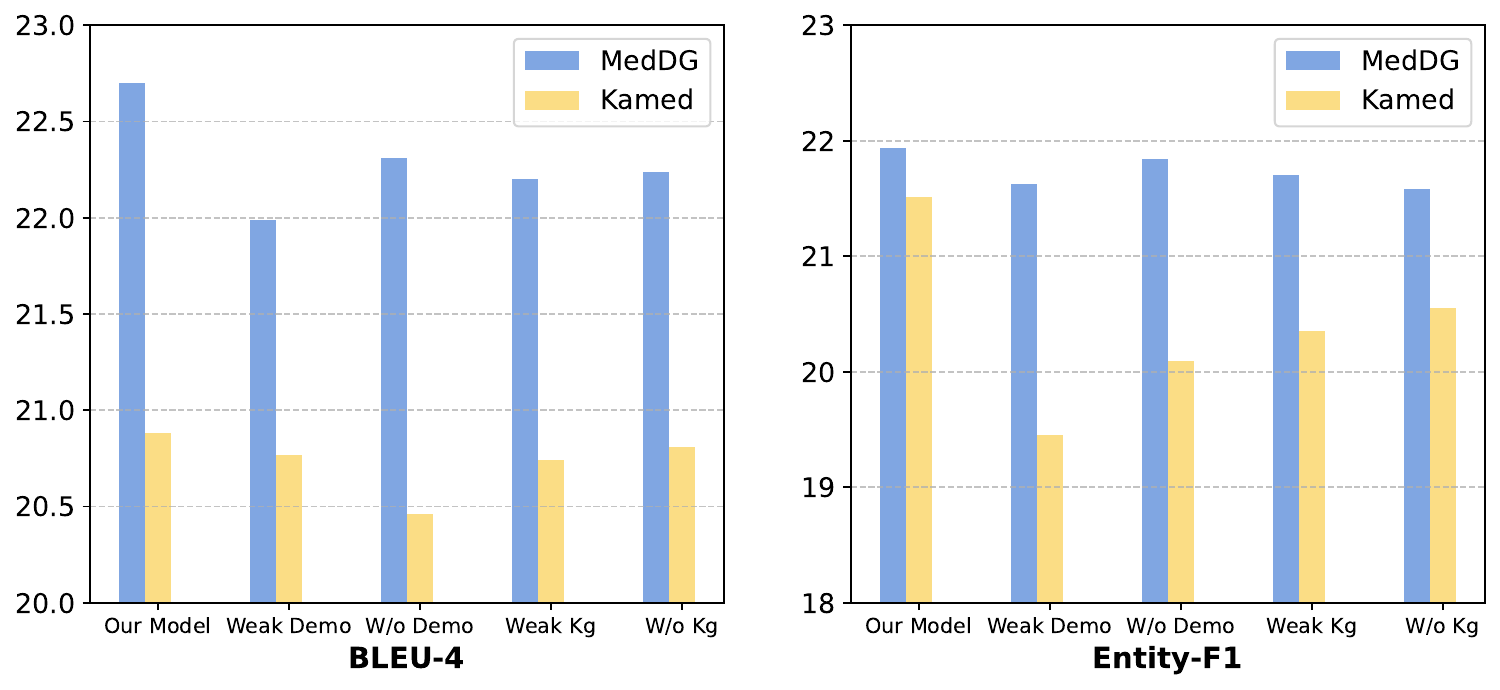}
    \caption{Comparison results of triplet filter and demo selector.}
    \label{fig:analysis}
\end{figure*}

\subsection{\mbox{Analysis of Triplet Filter and Demo Selector}}

To further verify the effectiveness of our triplet filter and demo selector modules, we introduce two new model variants: 
(1) \textbf{Weak Kg}: Instead of entirely removing the knowledge triplets from the prompt (w/o Kg), this variant bypasses the filtering rule and directly retrieves the triplets connected to entities in the most recent utterance from the knowledge graph, randomly selecting $M$ triplets from the one-hop connections. 
(2) \textbf{Weak Demo}: In this variant, demonstration examples are selected randomly, without any alignment process to ensure relevance.

The results in Figure \ref{fig:analysis} show that 
both variants exhibit significant performance drops across key metrics. Notably, we observe that merely increasing the quantity of knowledge triplets, without applying the triplet filter, harms the model’s performance. This suggests that indiscriminate use of knowledge can introduce noise, overwhelming the model and reducing its ability to generate accurate responses. Similarly, the random selection of demonstrations also leads to a decline in generation quality, highlighting the importance of the demo selector’s alignment process. These findings confirm that both the triplet filter and demo selector are essential for improving the accuracy and relevance of the generated medical dialogues.

\subsection{Case Study}\label{sec:case study}

\begin{figure*}[ht]
    \centering
    \includegraphics[width=\textwidth]{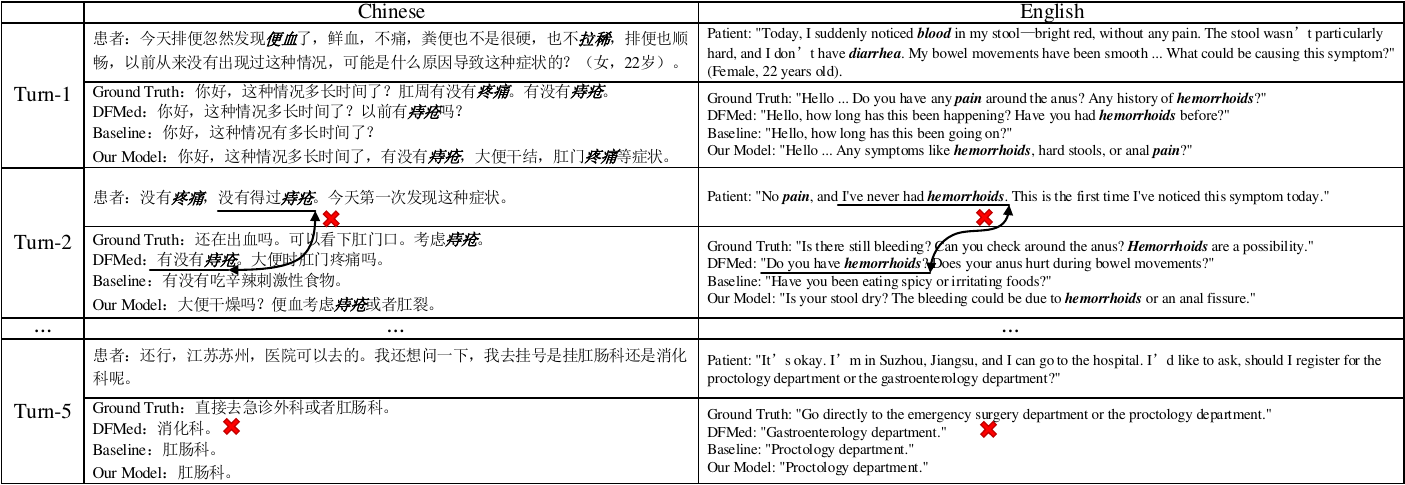}
    \caption{A running case comparing \model with baselines, highlighting that \model predicts more accurate medical entities and generates more relevant responses.}
    
    \label{fig:case}
\end{figure*}

Figure \ref{fig:case} illustrates a running example from the MedDG dataset, showcasing the dialogue across multiple turns. 

In Turn-1, \model demonstrates its ability to focus on the key entities ``hemorrhoids'' and ``pain'', producing a response that closely matches the Ground Truth. In comparison, both DFMed and the Baseline models fail to fully capture these entities, leading to incomplete responses. This highlights \model's superior entity prediction capabilities and its ability to generate more comprehensive inquiries that better address patient concerns.

In Turn-2, DFMed correctly predicts the entity ``hemorrhoids'' but ignores the patient's earlier statement ``I've never had hemorrhoids'', leading to a contradictory response. In contrast, \model remains consistent with the patient's current health information, thus leading to a more accurate and contextually appropriate diagnosis.

In Turn-5, \model still manages to provide a relevant and informative response. This is largely due to its effective use of the retrieved knowledge and its ability to infer information from the overall context. This example further demonstrates \model's robustness, showcasing its ability to handle situations where explicit entity cues are absent, yet still deliver meaningful and accurate dialogue.

Overall, these cases emphasize the advantages of \model in not only predicting relevant medical entities but also in maintaining contextual coherence throughout the conversation, leading to more reliable and patient-centered interactions. This illustrates how \model surpasses existing baselines, which often struggle with maintaining context consistency and addressing patient concerns comprehensively.

\subsection{Human Evaluation}
In addition to automatic evaluation, we conduct human evaluation experiments with a dedicated team. The volunteers are all medical doctoral and master's students with extensive experience in annotating medical conversations, who have been working on related projects for the past few years, and can ensure the reliability of the correct judgments.
The evaluators are tasked with scoring the responses and rating three aforementioned metrics (FLU, KC, OQ) using a scale from 1 (poor) to 5 (excellent).

As shown in Table \ref{Human Ablation}, \model consistently outperforms other baseline models across all three metrics. Notably, the scores for \model are the closest to the ground-truth responses, suggesting its higher level of alignment with expert expectations. This reinforces the idea that \model's specialized design, particularly the integration of entity-aware mechanisms and dynamic prompt adjustment, leads to more reliable and contextually relevant responses.

A key insight from this evaluation is that our framework's prompt design and dynamic adjustments significantly enhance the generation quality of large language models (LLMs). The results indicate that simply fine-tuning LLMs with generic prompts is insufficient for the complex nature of MDS. In contrast, \model leverages tailored prompt strategies and knowledge refinement, allowing it to generate responses that are not only more fluent but also exhibit higher medical accuracy. These findings highlight the advantage of our system, demonstrating that the combination of entity prediction, knowledge refining, and context-aware prompts enables the generation of higher-quality medical dialogues compared to simple fine-tuning strategies.

\begin{table}[ht]
    \centering
    \caption{Comparison results for human evaluation. Each metric ranges from 1 to 5.}
    \label{Human Ablation}
    \begin{tabular}{llll}
    \hline
        \textbf{Method} & \textbf{FLU} & \textbf{KC} & \textbf{OQ} \\ \hline
        Ground-truth & \textbf{3.70} & \textbf{3.75} & \textbf{3.95} \\ \hline
        DFMed & 3.42 & 3.57 & 3.65 \\ 
        E-A\&Cxt only & 2.91 & 3.05 & 3.14 \\ 
        \model & \underline{3.55} & \underline{3.68} & \underline{3.79} \\ \hline
    \end{tabular}
\end{table}

\section{Conclusion}
In this paper, we propose \fullmodel (\model). We introduce a variational knowledge refining mechanism for more accurate medical entity predictions and knowledge-driven responses.
We also develop a dynamic prompt adjustment method that adapts system prompts in real-time to the patient's evolving condition, ensuring more personalized and contextually relevant multi-turn medical dialogue generation.
Extensive experiments on two benchmarks verify that \model can achieve the best performance in terms of both text generation and medical entity-based metrics. These findings underscore \model's potential to improve the quality and reliability of MDS, paving the way for more context-aware and medically sound interactions in healthcare settings.

\section*{Limitations}
While our model achieves state-of-the-art performance in medical dialogue generation, two key limitations present opportunities for future improvement: (1) Unlike textual medical knowledge, cross-modal knowledge data has not been fully explored to enhance the capture of patient conditions. (2) The emotional support capabilities of current MDS are still passive rather than active. Appropriate comforting strategies are needed while maintaining medical accuracy.

\section*{Ethical Considerations}
The development and deployment of the medical dialogue system prioritize user safety, privacy, and the responsible use of AI in healthcare. All data used for training is anonymized. The proposed system is clarified to be intended as an assistive tool, not a replacement for professional medical advice, and should be used in conjunction with consultation from qualified healthcare providers.

\section*{Acknowledgments}

This work was supported by the Beijing
Outstanding Young Scientist Program NO. BJJWZYJH012019100020098, and Intelligent Social Governance Platform, Major Innovation \& Planning Interdisciplinary Platform for the ``Double-First Class'' Initiative, Renmin University of China, the Fundamental Research Funds for the Central Universities, and Public Computing Cloud, Renmin University of China, the fund for building world-class universities (disciplines) of Renmin University of China.

% Bibliography entries for the entire Anthology, followed by custom entries
%\bibliography{anthology,custom}
% Custom bibliography entries only
\bibliography{medref}

\end{document}